\documentclass[letterpaper,10pt,journal,twoside]{IEEEtran}

\usepackage[T1]{fontenc}
\usepackage{xcolor}
\usepackage{amsmath,amsfonts,amssymb}
\usepackage{mathtools}
\usepackage{graphicx}
\usepackage{pgfplots}
\usepackage{tikz}
\usetikzlibrary{arrows.meta}
\usepackage{pifont}
\usepackage[font=footnotesize]{caption}

\usepackage{cite}
\usepackage{hyperref}
\usepackage{cleveref}
\newcommand{\st}{\text{s.t.}\ }
\newcommand{\trans}{{\mkern-1.5mu\mathsf{T}}}
\newcommand{\myparagraph}[1]{\vspace{3pt} \noindent \textbf{#1} \ }
\newcommand{\cmark}{\ding{51}} 
\newcommand{\xmark}{\ding{55}} 

\crefname{section}{Sec.}{Secs.}
\Crefname{section}{Sec.}{Secs.}
\crefname{figure}{Fig.}{Figs.}
\Crefname{figure}{Fig.}{Figs.}
\crefname{table}{Table}{Tables}
\Crefname{table}{Table}{Tables}
\crefname{algocf}{Alg.}{Algs.}
\Crefname{algocf}{Alg.}{Algs.}

\newcommand{\bricksim}{\textsc{BrickSim}}

\providecommand{\Vector}[1]{\boldsymbol{#1}}
\providecommand{\Matrix}[1]{\boldsymbol{\mathbf{#1}}}
\providecommand{\Set}[1]{\mathbb{#1}}

\DeclareMathOperator{\round}{round}
\DeclareMathOperator*{\argmin}{\arg\!\min}

\newcommand{\tf}[2]{{}^{#1}\Matrix{T}_{#2}}

\newcommand{\fullwidthlist}{%
\IEEEnocalcleftmargintrue%
\setlength{\leftmargin}{0pt}%
\setlength{\itemindent}{\parindent}%
\addtolength{\itemindent}{\labelwidth}%
\addtolength{\itemindent}{\labelsep}}

\begin{document}

\title{\bricksim{}: A Physics-Based Simulator\\ for Manipulating Interlocking Brick Assemblies}

\markboth{IEEE Robotics and Automation Letters. Preprint Version. Accepted December, 2026}%
{Wen \MakeLowercase{\textit{et al.}}: \bricksim{}: Simulating Interlocking Brick Assemblies}

\author{Haowei~Wen, Ruixuan~Liu, Weiyi~Piao, Siyu~Li, and Changliu~Liu%
\thanks{Manuscript received: March 18, 2026; Revised: October 19, 2026; Accepted: November 17, 2026.}%
\thanks{This paper was recommended for publication by Editor Ashis Banerjee upon evaluation of the Associate Editor and Reviewers' comments.
This material is based upon work supported in part by the National Science Foundation under Grant No.~2403061. \textit{(Corresponding author: Changliu Liu.)}}%
\thanks{The authors are with the Robotics Institute, Kenneth C. Griffin School of Computer Science, Carnegie Mellon University, Pittsburgh, PA 15213 USA
(e-mail: haoweiw@andrew.cmu.edu; ruixuanl@andrew.cmu.edu; wpiao@andrew.cmu.edu;
siyuli@andrew.cmu.edu; cliu6@andrew.cmu.edu).}}

\IEEEaftertitletext{%
\begin{center}
\includegraphics[width=\textwidth]{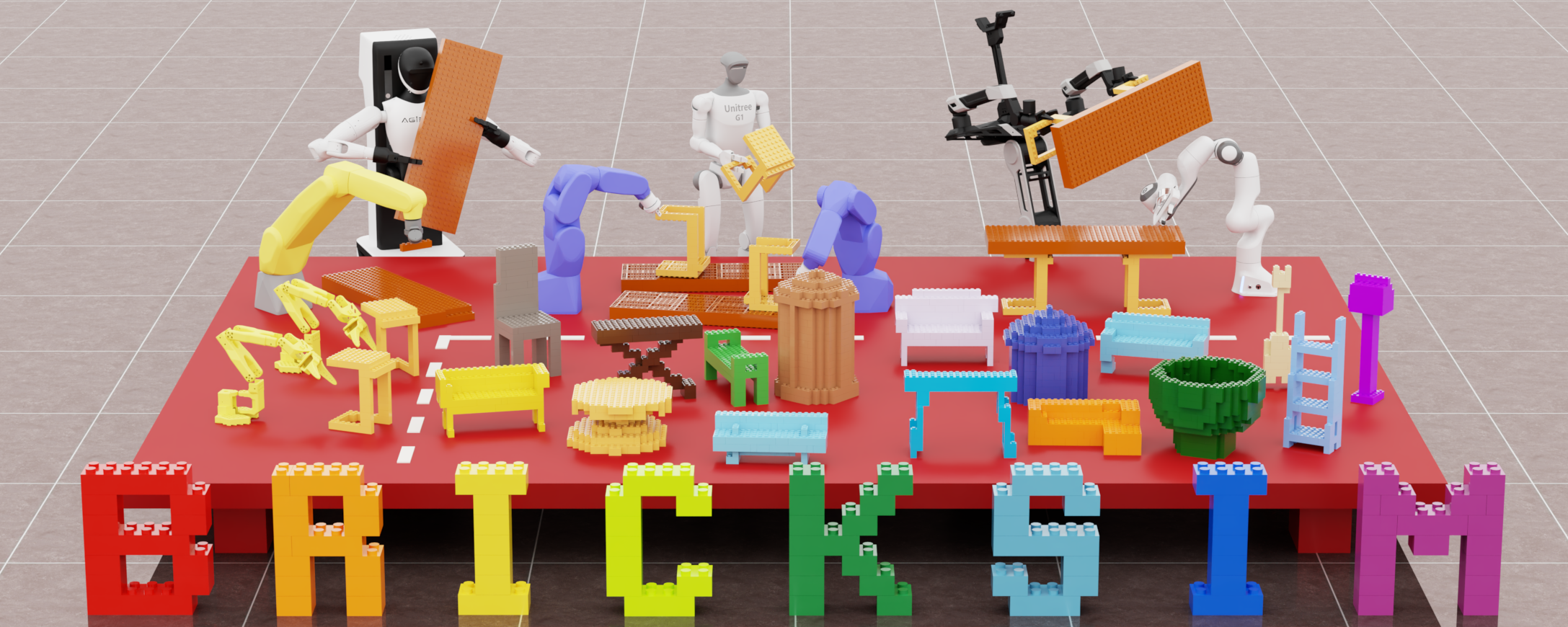}
\captionof{figure}{
    \textbf{\bricksim{} Overview.} 
    \bricksim{} enables high-fidelity simulation of complex brick assembly tasks and seamless integration with diverse robotic platforms. This teaser shows a representative multi-robot workflow for building a brick table alongside a gallery of diverse brick structures. All robotic and structural assets are natively instantiated within the \bricksim{} environment, arranged here for illustrative purposes.
}
\label{fig:teaser}
\end{center}%
}
\maketitle

\begin{abstract}

Interlocking brick assemblies provide a standardized yet challenging testbed for contact-rich and long-horizon robotic manipulation, but existing rigid-body simulators do not faithfully capture snap-fit mechanics.
We present \bricksim{}, the \textit{first} real-time physics-based simulator for interlocking brick assemblies.
\bricksim{} introduces a compact mechanics model for snap-fit connections, reduced to the micro-displacements of the bricks,
and solves the resulting internal force distribution by energy minimization.
Combined with a hybrid architecture that delegates rigid-body dynamics to the underlying physics engine while handling snap-fit mechanics separately,
\bricksim{} enables real-time, high-fidelity simulation of assembly, disassembly, and structural collapse.
On 250 real-world assemblies, \bricksim{} achieves 97.2\% accuracy in static stability prediction with an average solve time of 1.3\,ms.
In dynamic drop tests, it also closely reproduces real-world structural collapse and breakage patterns.
Built on Isaac Sim, \bricksim{} further supports seamless integration with a wide variety of robots and existing pipelines.
We demonstrate robotic construction of brick assemblies using \bricksim{},
highlighting its potential as a foundation for research in dexterous, long-horizon robotic manipulation.
\bricksim{} is open-source, and the code is available at
\url{https://github.com/intelligent-control-lab/BrickSim}.

\end{abstract}


\section{Introduction}\label{sec:intro}

\IEEEPARstart{B}{rick} assembly provides a compelling and widely adopted testbed for studying contact-rich manipulation \cite{popov2017dataefficient, liu2024lightweight, chen2023sequential, gu2025manualvla}, long-horizon planning \cite{nagele2020legobot, liu2024physics, huang2025apexmr}, and physical reasoning \cite{luo2015legolization, pletz2023brickfem, liu2024stablelego, pun2025generating}.
While the individual bricks are standardized, their combinatorial compositions create highly diverse assembly tasks.
These tasks are particularly challenging because they tightly couple local execution and global objectives.
Robots must precisely execute each snap-fit insertion and simultaneously plan ahead to achieve the final structural goal.

Physics simulators \cite{koenig2004gazebo,todorov2012mujoco,nvidiaisaacsim} are essential for studying such tasks at scale,
as they enable the large-scale data generation required for robot learning, safe and repeatable experimentation, and support for diverse robot embodiments.
Although existing simulators excel at modeling rigid-body dynamics,
they often fail to accurately capture interlocking contacts arising from micro-elastic deformation and friction,
such as the snap-fit connections between bricks.
In the absence of physically plausible simulators,
prior work on robotic brick manipulation has either focused on simplified block-stacking tasks \cite{groth2018shapestacks, bapst2019structured, fazeli2019seefeelact,goldberg2025bloxnet,xu2025stack}, in which smooth, non-interlocking blocks limit expressiveness and task complexity, or relied on real-world data collection \cite{liu2024lightweight,huang2025apexmr,gu2025manualvla}, which is costly and difficult to scale.
These limitations call for a simulator that preserves stable assemblies, reproduces collapse and breakage under disturbances, runs in real time, and integrates easily with modern robotic manipulation pipelines.

To address this gap, we present \bricksim{}, a real-time physics-based simulator for interlocking brick assemblies.
\bricksim{} augments a general-purpose physics engine with explicit modeling of snap-fit mechanics.
It detects valid stud-hole engagements to form new connections and breaks existing connections when they are overloaded.
The hybrid design preserves the efficiency of the underlying physics engine while enabling accurate simulation of snap-fit connections.
Built on Isaac Sim \cite{nvidiaisaacsim}, \bricksim{} can be seamlessly integrated into existing robotic manipulation pipelines.

\myparagraph{Contributions.}
Our contributions are as follows:
\begin{itemize}
\item We present \bricksim{}, the first real-time physics-based simulator for interlocking brick assemblies that supports physically realistic assembly, disassembly, and structural collapse, and integrates directly with Isaac Sim for robotic workflows.
\item We propose a hybrid simulation architecture that handles snap-fit mechanics separately from rigid-body dynamics, allowing us to leverage both the efficiency of the underlying rigid-body physics engine and the accuracy of explicit snap-fit modeling.
\item We develop a compact mechanics model for brick assemblies, reduced to the micro-displacements of the bricks, that yields a structured sparse convex energy minimization problem for internal force distribution, enabling real-time and accurate prediction of connection breakage.
\item We validate \bricksim{} across both static and dynamic scenarios, achieving 97.2\% accuracy in static stability prediction on 250 real-world assemblies, and closely reproducing structural collapse and breakage patterns in dynamic drop tests.
\end{itemize}

\section{Related Work}\label{sec:related_works}
We summarize a comparison of \bricksim{} with prior methods in \cref{table:simulator_comparison}, highlighting its ability to combine physically realistic modeling, temporal dynamics, efficient runtime, and seamless robot integration.

\begin{figure}
\centering
\includegraphics[width=\linewidth]{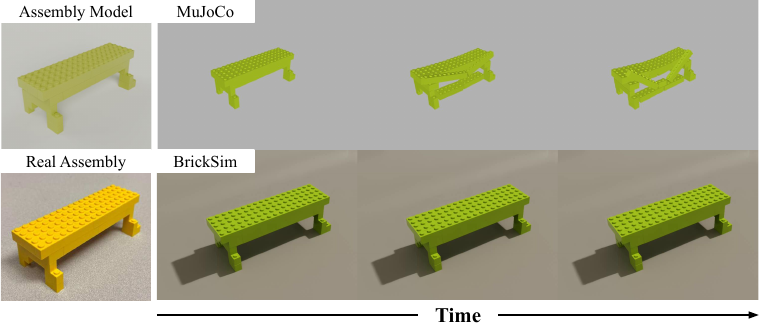}
\caption{
  \textbf{Comparison of the temporal behavior of a physically stable brick assembly in MuJoCo \cite{todorov2012mujoco} and \bricksim{}}.
  The real assembly remains intact. Starting from the same initial state, the assembly simulated in MuJoCo progressively collapses, while \bricksim{} preserves the structure, consistent with real-world behavior.
}
\label{fig:failure_case}
\end{figure}

\myparagraph{Brick Assembly Stability Analysis.}
Determining the physical stability of a brick assembly is crucial for designing brick structures.
Early work on brick assembly generation relied on heuristics to estimate constructability and optimize brick structures \cite{gower1998lego, ono2013legobuilder, testuz2013automatic}.
While practical, these heuristics do not provide a rigorous prediction of whether an assembly will stand or collapse.
Recent efforts have inferred stability by explicitly modeling equilibrium and frictional limits, yielding more physically grounded predictions \cite{luo2015legolization, kollsker2021legoconstructions, liu2024stablelego}.
This improved physical understanding has benefited subsequent research on robotic brick assembly \cite{pun2025generating, huang2025apexmr}.
Nevertheless, these methods focus primarily on static or quasi-static scenarios rather than the dynamic and temporal behaviors that arise after instability occurs.
They also explicitly parameterize loads using fine-grained local force variables at contact points.
Although this yields more physically grounded modeling, the resulting problem size is also larger, especially for multi-stud connections, making real-time analysis challenging.

\myparagraph{Physics-based Simulators.}
General-purpose physics engines \cite{koenig2004gazebo, todorov2012mujoco, nvidiaisaacsim} simulate rigid-body dynamics quite well, but they fail to capture the snap-fit mechanics arising from micro-elastic deformation and frictional forces.
Because of these modeling limitations,
physically stable brick structures often collapse spontaneously in standard rigid-body simulators,
as shown in \cref{fig:failure_case}.
While recent work augments rigid-body simulators with dedicated joining models for snap-fit assembly \cite{laemmle2022cabinetassembly},
it is tailored to a specific cabinet assembly task, rather than multi-body brick assemblies with many interlocking connections and dynamically evolving topology.
Closely related to our work, BrickFEM~\cite{pletz2023brickfem} uses finite element methods to predict the physical stability and temporal dynamics of brick assemblies.
While effective, its approach is computationally expensive, thus limiting its applicability to real-time simulation and large-scale assemblies.

\begin{table}[t]
\centering
\caption{
  \textbf{Comparison of representative methods for brick assembly analysis and simulation.}
  FSA: Force-based Stability Analysis \cite{luo2015legolization,liu2024stablelego}.
  Rigid-body simulators include Gazebo \cite{koenig2004gazebo}, MuJoCo \cite{todorov2012mujoco}, and Isaac Sim \cite{nvidiaisaacsim}.
}
\label{table:simulator_comparison}
\setlength{\tabcolsep}{4pt}
\resizebox{\linewidth}{!}{
\begin{tabular}{c  c  c c c} 
\hline
                   & Rigid-body  & FSA    & BrickFEM \cite{pletz2023brickfem} & \bricksim{} (Ours) \\ \hline
Snap-fit Mechanics & \xmark      & \cmark & \cmark                            & \cmark             \\
Temporal Dynamics  & \cmark      & \xmark & \cmark                            & \cmark             \\
Efficient Runtime  & \cmark      & \cmark & \xmark                            & \cmark             \\
Robot Integration  & \cmark      & \xmark & \xmark                            & \cmark             \\
\hline
\end{tabular}
}
\end{table}

\section{Overview of \bricksim{}}\label{sec:overview}

\begin{figure*}
\centering
\resizebox{\linewidth}{!}{\input{figs/BrickSim_Architecture.tikz}}
\caption{
    \textbf{\bricksim{} Architecture.}
    \bricksim{} is built on top of Isaac Sim and augments its simulation stack (blue).
    The \emph{Brick Topology Graph} (yellow) maintains a snap-fit connectivity graph between bricks and synchronizes it to the physics engine.
    The \emph{Assembly Monitor} (green) detects valid stud-hole engagements and adds new connections, while the \emph{Breakage Detector} (red) evaluates existing connections under external loads and removes those that fail.
    \bricksim{} provides a configurable simulation environment that supports diverse robotic platforms and a rich set of brick assets, including individual bricks and predefined assemblies. Bricks can be manipulated directly from the UI or by a robot controlled via teleoperation or custom policies.
}
\label{fig:system_architecture}
\end{figure*}

\myparagraph{Simulator Interface.}
\bricksim{} is a ready-to-use simulator for robotic manipulation of interlocking brick assemblies.
It inherits Isaac Sim's support for robotic platform integration, rigid-body simulation, and rendering.
As shown on the left of \cref{fig:system_architecture}, users can configure both robotic embodiments and brick assets, including individual bricks and predefined assemblies from the StableText2Brick dataset \cite{pun2025generating}.
The simulator supports direct UI interaction with bricks via drag-and-drop operations.
For robotic manipulation, users can either teleoperate the robot \cite{cadene2024lerobot} or deploy their own policies.
Video demonstrations are provided in the code repository.

\myparagraph{Core Architecture.}
As shown on the right of \cref{fig:system_architecture},
\bricksim{} augments the Isaac Sim backbone with three modules: the \emph{Brick Topology Graph} (BTG), the \emph{Assembly Monitor} (ASM), and the \emph{Breakage Detector} (BRD).
The BTG stores snap-fit connectivity, maintains consistent relative poses of bricks within each connected component, and synchronizes rigid constraints and collision filtering with the underlying PhysX physics engine.
The ASM monitors contact reports and creates new connections when two valid stud-hole engagements are detected between two bricks.
The BRD evaluates the current assembly under external loads, solves for the internal force distribution, and removes overloaded connections.
At each simulation step, Isaac Sim advances the rigid-body state and reports contact and impulse information.
The ASM and BRD then update the topology, and the updated BTG is synchronized back to PhysX for the next step.
\cref{sec:topology_graph} details the BTG representation and its integration with the physics engine,
\cref{sec:assembly_monitor} presents the criteria for connection formation, and
\cref{sec:breakage_detector} introduces our snap-fit mechanics model, solver pipeline, and breakage criterion.

\section{Brick Topology Graph} \label{sec:topology_graph}

We maintain a \textit{Brick Topology Graph} (BTG) to continuously track the state of the assembly, including the specifications of each brick and the connections between them. 
BTG decouples the topological relationships among bricks from their physical poses in the simulator, enabling us to mitigate numerical drift introduced by the physics solver. 
Moreover, it allows us to leverage efficient graph-based algorithms for connectivity updates and structural reasoning.

\myparagraph{BTG Representation.}
We represent all bricks in the scene as an undirected graph $G=(\Set{V},\Set{E})$,
where nodes $\Set{V}$ represent bricks and edges $\Set{E}$ represent kinematic connectivity.
Specifically, a single edge is defined between two bricks if they are physically joined by one or more snap-fit connections. 
Each connected \textit{component} in the graph corresponds to a rigidly connected assembly,
and all bricks within that component have fixed relative poses with respect to one another.
A rigid transform $\tf{u}{v} \in SE(3)$ is associated with each edge $\{u,v\}\in\Set{E}$,
representing the relative pose of brick $v$ in the frame of brick $u$, where $u, v\in \Set{V}$.
Traversing the same edge from $v$ to $u$ uses the inverse transform $\tf{v}{u}=\tf{u}{v}^{-1}$.
Starting from one brick, we can traverse the connected component to compute the relative pose of every brick in that component
by accumulating the transforms along the path.
A key invariant is that all paths between any two bricks in the same component must yield the same relative pose.
We enforce this whenever a new connection is added.

\begin{figure}
\centering
\includegraphics[width=\linewidth]{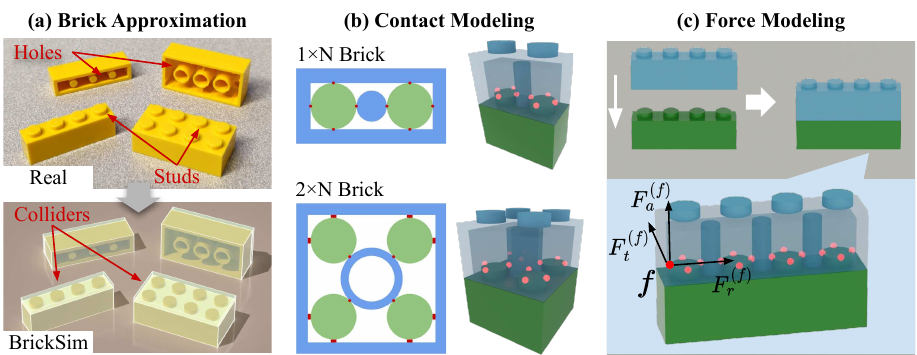}
\caption{
    \textbf{Brick and Connection Modeling.}
    (a) Brick geometries in the real world and the approximated collision models in \bricksim{}.
    (b) Each stud has 3 or 4 contact points with the walls or tubes of the upper brick, represented as red spheres.
    (c) Bricks are assembled by inserting the studs of the lower brick (green) into the holes of the upper brick (blue).
}
\label{fig:representations}
\end{figure}

\myparagraph{Connections.}
Each brick exposes a set of \textit{interfaces} that define admissible connection regions:
a \textit{stud} interface on the top surface and a \textit{hole} interface on the bottom surface for cuboid bricks, as shown in \cref{fig:representations}~(a).
Each interface is equipped with a local grid coordinate system aligned to brick units (unit length $L_U=8$\,mm) and a polarity (stud vs.\ hole).
A snap-fit \emph{connection} is defined between a stud interface $I_s^{(u)}$ on brick $u$ and a hole interface $I_h^{(v)}$ on brick $v$.
Each connection is parameterized by a discrete tuple $(\Vector{o},\psi)$,
where $\Vector{o} \in \Set{Z}^2$ is the planar grid offset in stud-grid units
and $\psi$ is the relative yaw quantized to multiples of $\frac{\pi}{2}$.
These discrete parameters induce a rigid transform between the two interface frames.

Multiple connections may exist between the same pair of bricks,
in which case all connections between them must induce the same brick-to-brick transform,
and we add only one edge to the topology graph to represent their connectivity.
While real $1 \times 1$ connections allow rotation about the vertical axis,
we do not allow this degree of freedom in \bricksim{} as a modeling simplification.
Supporting free yaw rotation for $1\times1$ connections is left for future work.

\myparagraph{Integration with Physics Engine.}
As illustrated in \cref{fig:representations}~(a), each brick is simulated as a rigid body with a simplified cuboid collision proxy to keep collision detection efficient.
Snap-fit behavior is introduced by explicitly checking for stud-hole engagement in \cref{sec:assembly_monitor} and adding rigid constraints.
For all bricks within the same connected component,
rigid constraints are added to fix their relative poses according to the BTG,
and collisions between them are disabled since the BRD in \cref{sec:breakage_detector} already models the forces between them.
Iterative solvers in physics engines can converge slowly when solving long chains of constraints.
Inspired by \cite{mueller2017longrangeconstraints},
we add extra constraints between random brick pairs within the same connected component,
sampled from a seeded random graph with maximum degree $d$,
to speed up convergence.

\section{Assembly Monitor} \label{sec:assembly_monitor}

Off-the-shelf rigid-body engines \cite{koenig2004gazebo,todorov2012mujoco,nvidiaisaacsim} do not model snap-fit mechanics.
To enable snap-fit assembly, we implement an \emph{Assembly Monitor} (ASM) that runs at every simulation step
to decide when two bricks should be connected based on contact reports.
At a high level, it
(1) enumerates candidate brick pairs from contacts,
(2) checks the geometric and force feasibility of each candidate connection, and
(3) commits the accepted connections to the BTG. 

\myparagraph{Candidate Generation.}
For each simulation step, we collect brick pairs that the physics engine reports to be in contact.
We then enumerate compatible stud-hole interface pairs as candidates.
For cuboid bricks, this corresponds to the stud interface on the top of the lower brick and a hole interface at the bottom of the upper brick.

\myparagraph{Geometric Criteria.}
A valid snap-fit connection must align with the discrete stud grid and have positive overlap.
Given a candidate stud interface $I_s^{(u)}$ on brick $u$ and hole interface $I_h^{(v)}$ on brick $v$,
we compute their relative pose:
\begin{equation}
\tf{I_s^{(u)}}{I_h^{(v)}} = \begin{bmatrix}
    \Matrix{R}              & \Vector{t} \\
    \Vector{0}^\trans    & 1
\end{bmatrix},
\quad
\Vector{t} = \begin{bmatrix} t_x & t_y & t_z \end{bmatrix}^\trans,
\label{eq:assembly_relative_pose}
\end{equation}
where $\Matrix{R} \in SO(3)$ is the rotation matrix and $\Vector{t} \in \Set{R}^3$ is the translation vector.
We then snap this relative pose to the nearest discrete configuration $(\Vector{o},\psi)$:
\begin{equation}
\Vector{o} = \round\Big( \frac{1}{L_U} \begin{bmatrix} t_x \\ t_y \end{bmatrix} \Big), \quad
\psi = \frac{\pi}{2} \round\left( \frac{\theta_\mathrm{yaw}}{\pi/2} \right),
\label{eq:assembly_snap}
\end{equation}
where $\theta_\mathrm{yaw}$ is the relative yaw angle extracted from $\Matrix{R}$.
We accept a candidate only if the snapped configuration is within a set of tunable geometric tolerances
$(\epsilon_z, \epsilon_\mathrm{tilt}, \epsilon_\psi, \epsilon_{xy})$:
\begin{itemize}
    \item \textbf{Vertical distance:} $|t_z - H_S| \le \epsilon_z$, where $H_S = 1.7$\,mm is the pre-engagement height implied by our collision model.
    \item \textbf{Tilt:} $\theta_\mathrm{tilt} \le \epsilon_\mathrm{tilt}$, where $\theta_\mathrm{tilt}$ is the angle between the two interface normals.
    \item \textbf{Yaw error:} $|\theta_\mathrm{yaw} - \psi| \le \epsilon_\psi$.
    \item \textbf{Planar error:} $\|[t_x,t_y]^\trans - L_U \Vector{o}\| \le \epsilon_{xy}$.
    \item \textbf{Positive overlap:} under $(\Vector{o},\psi)$, the stud grid of $I_s^{(u)}$ and the hole grid of $I_h^{(v)}$ overlap with positive area.
\end{itemize}

\myparagraph{Force Criteria.}
Geometric alignment alone is insufficient because a real snap-fit requires a pressing force to engage.
We therefore gate assembly using a minimum compressive load along the stud-engagement axis.
Concretely, we read the normal contact impulse between the two bricks over the simulation step from the physics engine
and convert it to an average force by dividing by the step duration $\Delta t$.
We then project this force onto the stud interface normal direction and require the resulting compressive force to exceed a fixed threshold $F_\mathrm{asm}$.

\myparagraph{Connection Creation.}
When a candidate satisfies both geometric and force criteria, the ASM emits a new connection $(I_s^{(u)},I_h^{(v)},\Vector{o},\psi)$.
The BTG then inserts the connection and updates the connected components as described in \cref{sec:topology_graph},
creating new rigid constraints and setting up collision filtering.
To improve stability during assembly, we snap the bricks to the exact discrete relative transform implied by $(\Vector{o},\psi)$.

\section{Breakage Detector} \label{sec:breakage_detector}

Real snap-fit connections can fail when overloaded, leading to disassembly or structural collapse.
We model this behavior with a \emph{Breakage Detector} (BRD) that evaluates each connected component of the BTG.
The BRD first constructs a mechanical model for each component from its topology.
At each simulation step, it computes the internal forces using equilibrium and energy minimization, and breaks connections that exceed frictional limits.

\subsection{Mechanical Model}

\myparagraph{Contacts.}
Contacts $\Set{C}$ arise from touching brick surfaces and transmit only compression.
The contact $c \in \Set{C}$ between bricks $i$ and $j$ applies frictionless normal force to them:
\begin{equation}
\Vector{F}_{i \leftarrow \ell} = -\Vector{F}_{j \leftarrow \ell} = -\lambda^{(\ell)} \Vector{\hat{n}}_c
\qquad \forall \ell \in \mathrm{Vtx}(\Omega_c),
\label{eq:breakage_contact_force}
\end{equation}
where $\Omega_c$ is the polygonal overlap between the two surfaces,
$\mathrm{Vtx}(\Omega_c)$ is the set of its vertices,
$\lambda^{(\ell)} \geq 0$ is the force magnitude at vertex $\ell$,
and $\Vector{\hat{n}}_c$ is the unit normal pointing from $i$ to $j$.

\myparagraph{Connections.}
Connections $\Set{K}$ arise from stud--hole engagements and carry tension, shear, and torsion.
The connection $k \in \Set{K}$ between bricks $i$ and $j$ includes an array of studs on brick $i$.
As illustrated in \cref{fig:representations}~(b), each stud touches the surrounding walls or tubes of brick $j$ at 3 or 4 contact points \cite{luo2015legolization,liu2024stablelego}.
Let $\Set{F}_k$ denote the set of all contact points of connection $k$,
$\Vector{\hat{n}}_k$ the unit vector along the stud axis pointing from $i$ to $j$,
$\Vector{n}_f$ the inward radial normal at contact point $f \in \Set{F}_k$,
and $\Vector{t}_f = \Vector{\hat{n}}_k \times \Vector{n}_f$ the tangential direction.
As shown in \cref{fig:representations}~(c), the force on brick $i$ at $f$ has
axial, radial, and tangential components $F_a^{(f)}$, $F_r^{(f)}$, and $F_t^{(f)}$:
\begin{equation}
\Vector{F}_{i \leftarrow f} = -\Vector{F}_{j \leftarrow f} = F_a^{(f)} \Vector{\hat{n}}_k + F_r^{(f)} \Vector{n}_f + F_t^{(f)} \Vector{t}_f.
\label{eq:breakage_connection_force}
\end{equation}
$F_r^{(f)}$ is the nonnegative radial force exerted by the wall or tube on the stud
and equals the preload $F_0^{(f)}$ in the unloaded reference state.
$F_0^{(f)}$ is computed once per topology by minimizing the radial elastic energy
of the entire connected component without gravity or other external loads,
during which we assume equal initial radial interference and use a common nominal preload $F_0$.
Because axial compression is already carried by contacts,
the axial force is tension-only, $F_a^{(f)} \geq 0$.
The axial and tangential forces share the friction limit.
We approximate the friction cone with a linear friction pyramid,
$F_a^{(f)} + |F_t^{(f)}| \leq \mu F_r^{(f)}$, where $\mu$ is the static friction coefficient.

\myparagraph{Rigid Micro-Displacements.}
We treat each brick as rigid and represent local elastic deformation with compliances.
Such deformation follows from small rigid displacements of the bricks.
We describe the micro-displacement of brick $i$ from the unloaded reference state by
the generalized coordinates $\Vector{q}_i=(\Vector{d}_i,\Vector{\theta}_i)$,
where $\Vector{d}_i$ is the translation of the brick's center of mass (COM) $\Vector{c}_i$
and $\Vector{\theta}_i$ is its small rotation vector.
Stacking all $\Vector{q}_i$ gives the decision variables $\Vector{q}$ of the BRD,
with 6 variables per brick, independent of the number of contact points or connections.
The displacement of brick $j$ relative to brick $i$ at point $\Vector{x}$ is
\begin{equation}
\Vector{\Delta}_{ij}(\Vector{x}) =
\Vector{d}_j + \Vector{\theta}_j \times (\Vector{x}-\Vector{c}_j)
-\Vector{d}_i - \Vector{\theta}_i \times (\Vector{x}-\Vector{c}_i).
\label{eq:breakage_relative_displacement}
\end{equation}
At contact point $f$ of connection $k$, located at $\Vector{x}_f$,
projecting $\Vector{\Delta}_{ij}(\Vector{x}_f)$ onto
$\Vector{\hat{n}}_k$, $\Vector{n}_f$, and $\Vector{t}_f$
gives the axial, radial, and tangential relative displacements
$e_a^{(f)}$, $e_r^{(f)}$, and $e_t^{(f)}$.
At vertex $\ell$ of contact $c$, located at $\Vector{x}_\ell$,
projecting $\Vector{\Delta}_{ij}(\Vector{x}_\ell)$ onto $\Vector{\hat{n}}_c$
gives the gap change $g^{(\ell)}$.
Here $e_r^{(f)}>0$ increases radial compression,
and $g^{(\ell)}>0$ indicates separation of the contact.
The radial and contact forces follow elastic laws:
\begin{equation}
F_r^{(f)} = [F_0^{(f)}+e_r^{(f)} / w_r]_+, \quad
\lambda^{(\ell)} = [-g^{(\ell)}]_+ / w_c,
\label{eq:breakage_normal_response}
\end{equation}
where $[\cdot]_+=\max(\cdot,0)$,
and $w_r,w_c>0$ are the radial and contact compliances.
The axial and tangential forces are the elastic forces projected onto the friction pyramid:
\begin{equation}
\begin{split}
(F_a^{(f)}, F_t^{(f)}) = \argmin_{\substack{F_a \geq 0 \\ F_a + |F_t| \leq \mu F_r^{(f)}}}
\ & w_a \left( F_a - \frac{e_a^{(f)}}{w_a} \right)^2 \\
& + w_t \left( F_t - \frac{e_t^{(f)}}{w_t} \right)^2,
\end{split}
\label{eq:breakage_friction_response}
\end{equation}
where $w_a,w_t>0$ are the axial and tangential compliances.
A contact point sticks when these forces lie inside the pyramid and slips when they lie on its boundary.
Frictional slip is confined to the axial--tangential plane and has no radial component,
corresponding to a non-associated flow rule.

\subsection{Force Distribution Optimization}
The BRD solves for internal forces to identify overloaded connections and select the ones to break.
A load is supportable if and only if there exists an internal force distribution that balances it within friction limits.
When no such force distribution exists, the constrained solve is infeasible and does not directly quantify overload at individual connections.
We therefore seek the smallest increases in the connection friction limits that admit equilibrium.
For each connection $k$, we introduce a dimensionless relaxation $v_k\geq0$,
raising the friction limit at every point $f\in\Set{F}_k$
to $\mu(F_r^{(f)} + F_0 v_k)$.
We minimize these relaxations before recovering the force distribution under the resulting limits.

\myparagraph{Contact Potential.}
The total contact potential $U$ is the sum of the local potentials of connections and contacts.
Per the non-associated friction law, we hold the normal force in friction limits fixed at $\bar{F}_r^{(f)}$,
so the optimization does not introduce radial slip.
Integrating \cref{eq:breakage_normal_response,eq:breakage_friction_response} under relaxed friction limits gives
\begin{align}
&U(\Vector{q},\Vector{v};\bar{\Vector{F}}_r)
=\sum_{k\in\Set{K}}\sum_{f\in\Set{F}_k}
\bigg[\frac{w_r}{2}\big(F_r^{(f)}\big)^2
+\max_{\makebox[0pt]{$\scriptstyle\substack{
F_a\geq0\\
F_a+|F_t|\leq\mu(\bar{F}_r^{(f)}+F_0v_k)}$}}
\Big(F_a e_a^{(f)}
\label{eq:breakage_potential}\\
&\;+F_t e_t^{(f)}
-\frac{w_aF_a^2+w_tF_t^2}{2}\Big)\bigg]
+\sum_{c\in\Set{C}}\sum_{\ell\in\mathrm{Vtx}(\Omega_c)}
\frac{w_c}{2}\big(\lambda^{(\ell)}\big)^2,
\notag
\end{align}
where $\Vector{v} = [v_k]_{k\in\Set{K}}$ stacks connection relaxations
and $\bar{\Vector{F}}_r = [\bar{F}_r^{(f)}]_{f\in\Set{F}_k, k\in\Set{K}}$ stacks the frozen normal forces.

\myparagraph{Equilibrium.}
The internal wrench required on each brick is computed from the physics engine's outputs:
\begin{equation}
\Vector{b}_i=\frac{1}{\Delta t}
\left(
\begin{bmatrix}
m_i(\Vector{v}_i^+-\Vector{v}_i^-)\\
\Matrix{I}_i^+\Vector{\omega}_i^+ - \Matrix{I}_i^-\Vector{\omega}_i^-
\end{bmatrix} -
\begin{bmatrix}
\Vector{J}_i^{\mathrm{ext}}\\
\Vector{H}_i^{\mathrm{ext}}
\end{bmatrix}
\right),
\label{eq:breakage_equilibrium_per_brick}
\end{equation}
where $\Delta t$ is the step duration and $m_i$ is the brick mass.
$\Vector{v}_i^\pm$, $\Vector{\omega}_i^\pm$, and $\Matrix{I}_i^\pm$
are the pre- and post-step COM linear velocity, angular velocity, and inertia tensor in the component frame.
$\Vector{J}_i^{\mathrm{ext}}$ and $\Vector{H}_i^{\mathrm{ext}}$
are the external linear and angular impulses about the COM,
including gravity and contacts with objects outside the component.
By virtual work, $-\partial U / \partial \Vector{q}_i$ is the resultant internal wrench on brick $i$,
so equilibrium requires
\begin{equation}
-\frac{\partial U(\Vector{q},\Vector{v};\bar{\Vector{F}}_r)}
{\partial\Vector{q}}
=\Vector{b},
\label{eq:breakage_equilibrium_compact}
\end{equation}
where $\Vector{b}$ stacks $\Vector{b}_i$.
Because $U$ is convex in $\Vector{q}$, for fixed $\Vector{v}$ and $\bar{\Vector{F}}_r$
this is the condition for minimizing
$U(\Vector{q},\Vector{v};\bar{\Vector{F}}_r)+\Vector{b}^{\trans}\Vector{q}$.

\myparagraph{Lexicographic Optimization.}
At each simulation step, we initialize $\bar{F}_r^{(f)}=F_0^{(f)}$
and solve for the internal force distribution lexicographically as follows:
\begin{enumerate}[\fullwidthlist]

\item Find the closest balanced load $\Vector{b}^*$:
\begin{equation}
\Vector{b}^*=\argmin_{\Vector{y}}\|\Vector{y}-\Vector{b}\|^2
\enspace\st
\sum_{i}
\begin{bmatrix}
\Vector{y}_i^F\\
\Vector{c}_i\times\Vector{y}_i^F+\Vector{y}_i^\tau
\end{bmatrix}
=\Vector{0},
\label{eq:breakage_project}
\end{equation}
where the sum runs over all bricks in the current component,
and $\Vector{y}_i^F$ and $\Vector{y}_i^\tau$ are the force and the torque about $\Vector{c}_i$ of $\Vector{y}_i$.
Residuals from the physics engine can give $\Vector{b}$ a net force and torque.
This step removes them, so a solution is always feasible in the following stages.

\item Find the minimum relaxation $\Vector{v}^*$:
\begin{equation}
\Vector{v}^*=\argmin_{\Vector{v}\geq\Vector{0}}\|\Vector{v}\|^2
\quad\st\ \exists\Vector{q}: -\frac{\partial U}{\partial\Vector{q}}=\Vector{b}^*.
\label{eq:breakage_relax}
\end{equation}

\item Find the equilibrium under this relaxation:
\begin{equation}
\Vector{q}^*\in\argmin_{\Vector{q}}
\big[U(\Vector{q},\Vector{v}^*;\bar{\Vector{F}}_r)
+\Vector{b}^{*\trans}\Vector{q}\big].
\label{eq:breakage_equilibrium}
\end{equation}
\end{enumerate}
Stages~2 and~3 are repeated using fixed-point iteration to obtain a self-consistent equilibrium.
The friction limits depend on the radial forces recovered in Stage~3,
so after each pass we update $\bar{F}_r^{(f)}\leftarrow F_r^{(f)}$.
We stop when the absolute normal-force change falls below a prescribed tolerance,
or when the iteration limit is reached.

In our implementation, we solve Stages~2 and~3 jointly using sparse Newton
iterations on a penalized formulation, then recover the lexicographic solution
by asymptotically expanding the optimality system on the identified active set
as the penalty parameter tends to infinity.

\subsection{Breakage Selection}
After solving the aforementioned problem, we compute a per-connection utilization score $u_k$:
\begin{equation}
u_k = \max_{f \in \Set{F}_k} \frac{ |F_t^{(f)}| + F_a^{(f)} }{ \mu F_r^{(f)} },
\label{eq:breakage_utilization}
\end{equation}
where $u_k > 1$ indicates that the connection is overloaded and likely to break.
The stability score is reported as the largest $u_k$ among all connection.
A score larger than 1 means unstable, and a score close to 1 means that its most loaded connection is close to breaking.

Although multiple connections may satisfy $u_k>1$,
breaking all of them at once would be overly aggressive and non-physical.
We therefore identify the most offending connection
and break the smallest number of overloaded connections containing it
whose removal disconnects the connected component.
Applying this procedure at every simulation step mimics the progressive breakage behavior observed in the real world.

\section{Results} \label{sec:experiment}

We evaluate \bricksim{} through static stability tests, dynamic drop tests, and robotic manipulation tasks.
Unless otherwise specified,
all experiments are conducted on a machine with an Intel i9-12900H CPU, 64 GB of RAM, and an NVIDIA RTX 3080 Ti Mobile GPU.
The following parameters are used:
$\epsilon_z = 1.0$\,mm,
$\epsilon_\mathrm{tilt} = 5^\circ$,
$\epsilon_\psi = 5^\circ$,
$\epsilon_{xy} = 2.0$\,mm,
$F_\mathrm{asm} = 1.0$\,N,
$d = 4$,
$\mu=0.2$, $\mu F_0=0.7$\,N~\cite{luo2015legolization},
and $w_a:w_r:w_t:w_c=1:1:1:0.1$.
The compliance scale $w_a=6.7\times10^{-5}$\,m/N is fit to the heights of a real staircase of $2\times4$ bricks,
each offset by two studs, measured after each of its 21 bricks.
The fixed-point tolerance is $10^{-4}F_0$, with at most 10 passes.
Newton uses a normalized residual tolerance of $10^{-10}$,
with at most 100 iterations.

\subsection{Static Stability Evaluation}

\begin{figure}[t]
\begingroup
\centering
\captionof{table}{
    \textbf{Comparison of Static Structural Stability Analysis.}
    Accuracy counts unsolved ones as incorrect. Solve time is computed over solved ones.
}
\label{table:static}
\setlength{\tabcolsep}{5pt}
\resizebox{\linewidth}{!}{
\begin{tabular}{cccc} 
\hline
                         & BrickFEM \cite{pletz2023brickfem} & StableLego \cite{liu2024stablelego} & BrickSim              \\ \hline
Solved (of 250)          & $129$                             & $\mathbf{250}$                      & $\mathbf{250}$        \\
Stable Correct (of 232)  & $128$                             & $205$                               & $\mathbf{232}$        \\
Unstable Correct (of 18) & $0$                               & $\mathbf{16}$                       & $11$                  \\ \hline
Accuracy                 & $51.2\%$                          & $88.4\%$                            & $\boldsymbol{97.2\%}$ \\
F1 (Unstable)            & $0.000$                           & $0.525$                             & $\mathbf{0.759}$      \\ \hline
Avg. Solve Time (s)      & $637.9$                           & $0.037$                             & $\mathbf{0.0013}$     \\ \hline
\end{tabular}
}
\vspace{10pt}
\endgroup

\centering
\includegraphics{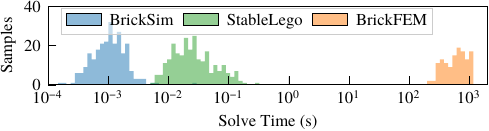}
\captionof{figure}{
    \textbf{Comparison of Solve Time.}
    Distributions of computation times for static structural stability analysis over all solvable structures.
}
\label{fig:time}
\end{figure}

\begin{figure}[t]
\centering
\includegraphics{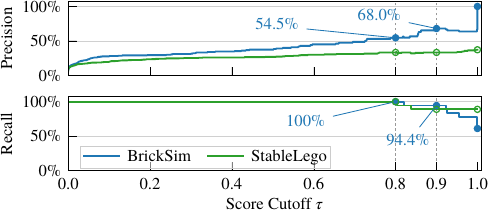}
\caption{
    \textbf{Conservative Stability Prediction.}
    Precision and recall for detecting the 18 unstable assemblies in \cref{table:static}.
    An assembly is flagged when its stability score is at least $\tau$,
    so $\tau=1$ gives the verdicts in \cref{table:static}.
}
\label{fig:stability_cutoff}
\end{figure}

We compare the static structural stability predicted by \bricksim{} with BrickFEM \cite{pletz2023brickfem} and StableLego \cite{liu2024stablelego}.
The evaluation set consists of 250 assemblies of up to 50 bricks from StableText2Brick~\cite{pun2025generating}.
Ground truth is established by physically building each assembly and observing whether it collapses without external support.
We set a 20-minute time limit for each solve.
Because BrickFEM itself does not give a stability verdict,
we consider an assembly as unstable if any node moves by more than 1\,mm during simulation.

\begin{figure}[t]
\centering
\includegraphics[width=\linewidth]{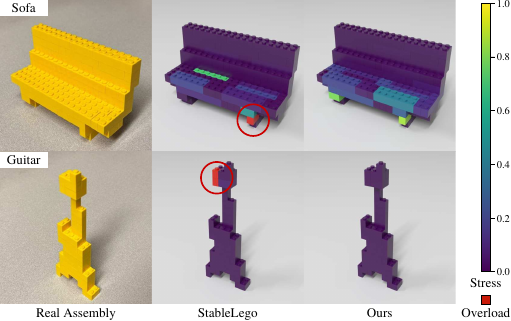}
\caption{
    \textbf{Comparison of Predicted Force Distributions.}
    Stress values range from 0 to 1, where lower values indicate smaller internal stress and higher values indicate greater stress, suggesting a higher risk of structural failure.
    Red: predicted collapsing bricks.
    StableLego predicts two bricks as overloaded, while they remain standing in the real world.
    \bricksim{} correctly classifies both as stable.
}
\label{fig:force_distribution_comparison}
\end{figure}

\bricksim{} achieves the highest accuracy and F1 score, as shown in \cref{table:static}.
It correctly classifies all 232 stable assemblies but misses 7 unstable ones.
StableLego misses 2 unstable assemblies and incorrectly flags 27 stable assemblies as unstable.
\cref{fig:force_distribution_comparison} illustrates two of these false-unstable predictions.
BrickFEM completes only 129 of the 250 analyses within the time limit.
It classifies every completed assembly as stable, mostly because the default simulation duration 0.5\,ms is too short for gravity to produce noticeable displacement, while a long duration that could show a collapse is computationally unaffordable.

The 7 unstable assemblies missed by \bricksim{} have
stability scores close to the failure threshold.
For applications that prioritize detecting unstable assemblies,
such as text-to-brick generation and assembly planning,
we evaluate more conservative stability prediction with lower score cutoffs.
\cref{fig:stability_cutoff} shows the resulting precision and recall.
At $\tau=0.8$, both methods identify all 18 unstable assemblies.
\bricksim{} incorrectly flags 15 stable assemblies, compared with 36 for StableLego.
Thus, at this point, \bricksim{} achieves the same recall while rejecting fewer stable assemblies.

In addition to physical correctness, we compare the solve times in \cref{table:static,fig:time}.
\bricksim{} achieves the lowest average solve time, significantly outperforming the baselines.
By reducing computation time to the milliseconds, \bricksim{} makes it feasible to run stability checks in real-time simulation.

\subsection{Dynamic Structure Simulation}

\begin{figure}[t]
\centering
\includegraphics[width=\linewidth]{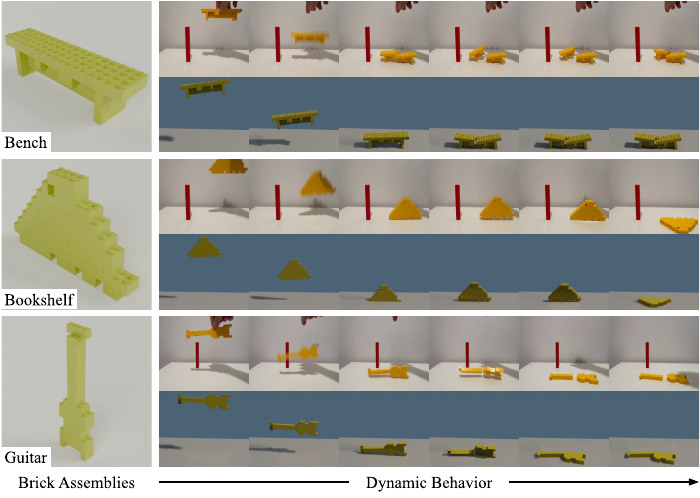}
\captionof{figure}{
    \textbf{Comparison of Temporal Dynamics.} Real (top row) and simulated (bottom row) behaviors of three brick assemblies dropped from midair.
}
\label{fig:dynamic}
\vspace{10pt}

\centering
\includegraphics[width=\linewidth]{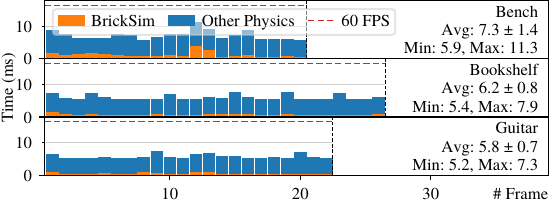}
\captionof{figure}{
    \textbf{End-to-End Frame Time in Drop Tests.}
    We measure the per-frame computation time for the three drop test simulations.
    Each stacked bar corresponds to one simulation frame and is decomposed into time spent in \bricksim{} and ``Other Physics'' (the remaining physics computation handled by Isaac Sim).
    The red dashed line marks the 60 FPS budget of 16.7 ms.
    Annotations report total frame times (avg $\pm$ std, min, max).
}
\label{fig:frame_time}
\end{figure}

Beyond static structures, we evaluate the fidelity of \bricksim{} in capturing temporal dynamics and structural failure.
Specifically, we perform controlled drop tests in which brick assemblies are released from a height of approximately 15 bricks (indicated by the red stick in \cref{fig:dynamic}) and impact a rigid table.
We compare the real-world behaviors against the dynamics simulated by \bricksim{}.
Upon impact, the real-world bookshelf remains intact, and \bricksim{} successfully simulates its strong interlocking connections.
In contrast, the bench and guitar split into two pieces upon collision.
\bricksim{} correctly simulates their structural collapse, despite minor differences in the motions of the split components after breakage.
Moreover, the specific breakage locations predicted by \bricksim{} match the real-world results.
These results show the physical fidelity of \bricksim{} in modeling the dynamic behaviors of brick assemblies.
The simulation timestep $\Delta t = 1/60\,$s.

Additionally, we measure the end-to-end frame times in the drop test simulations, as shown in \cref{fig:frame_time}.
Across all three experiments, the simulation strictly adheres to the 16.7 ms budget required for 60 FPS.
While frame times briefly peak during breakage due to BTG updates, they still remain below the real-time threshold, and the vast majority of frames process in under 10 ms.
These results demonstrate that \bricksim{} successfully maintains real-time performance throughout the simulation.

\subsection{Robotic Manipulation}

\begin{figure}[t]
\centering
\begingroup
\captionof{table}{
    \textbf{Robotic Brick Assembly Results.}
    Results over 200 trials, each with a random initial brick pose and target connection.
    Completion time (mean $\pm$ std) is computed over successful trials only.
}
\label{table:robot}
\footnotesize
\begin{tabular*}{\linewidth}{@{\extracolsep{\fill}}ccc}
\hline
                    & Scripted Expert & DAgger~\cite{ross2011dagger} \\ \hline
Success Rate        & $92.5\%$        & $39.5\%$                     \\
Completion Time (s) & $3.4 \pm 1.1$   & $4.1 \pm 1.5$                \\ \hline
\end{tabular*}
\endgroup

\medskip
\includegraphics[width=\linewidth]{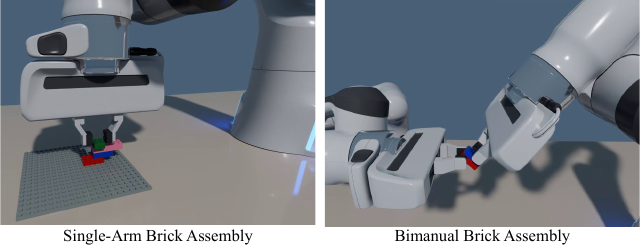}
\caption{
    \textbf{Robotic Brick Manipulation using \bricksim{}.}
    Left: a Franka robot is building an assembly by sequentially stacking bricks onto a baseplate on the table.
    Right: two Franka robots are cooperating to assemble two bricks in hand. One robot is holding the red brick, while the other robot is inserting the blue brick into it.
}
\label{fig:robot_demo}
\end{figure}

We further demonstrate robot integration in \cref{fig:robot_demo}.
Both demonstrations are implemented using the Python API provided by \bricksim{}
and leverage Isaac Sim's built-in robotic assets and motion-planning pipelines.
We also evaluate robot policies quantitatively on a single brick assembly task in Isaac Lab~\cite{mittal2025isaaclab}, in which a Franka robot picks up a $2\times4$ brick at a random pose on the table and assembles it onto a baseplate at commanded target position.
We compare a scripted expert with an MLP policy trained with DAgger~\cite{ross2011dagger} in \cref{table:robot}.
The DAgger policy is first trained on 1,000 successful expert trajectories and then on 50 DAgger rounds. Behavior cloning on the same demonstrations succeeds in none of the trials.
Demonstration videos are provided in the code repository.

\suppressfloats[t]

\section{Discussion} \label{sec:discussion}

\bricksim{} brings brick assembly into simulation, thus promoting reproducibility, large-scale benchmarking, and rapid prototyping of algorithms in robot learning and manipulation.
We hope \bricksim{} will foster broader collaboration across the robotics, graphics, and physics simulation communities and serve as a stepping stone toward more generalizable solutions for contact-rich, long-horizon manipulation and real-world assembly tasks.

\myparagraph{Limitations.}
First, \bricksim{} models assembly as a discrete process, so progressive insertion or partially inserted bricks cannot be simulated. A more detailed simulation of the insertion process is left for future work.
Second, our current robot evaluation covers only single-brick assembly. We plan to build benchmarks with more complex brick assembly tasks in the future.
Third, \bricksim{} currently supports parts with fixed snap-fit connections, such as
bricks, plates, and slopes. We aim to expand support to functional parts such as gears and wheels, whose degrees of freedom allow more complex interactions and dynamic behaviors.

\section*{Acknowledgment}
The authors thank Maggie Cai, Yuanliu Huang, and Bowei Li for assisting with experiments.
The first author gratefully acknowledges Prof. Zhenlong Du of Nanjing Tech University for early guidance on ideas that informed this work.

\bibliographystyle{IEEEtran}
\bibliography{references}

\end{document}